\documentclass{article}

\usepackage[final]{neurips_2024}

\usepackage[utf8]{inputenc} 
\usepackage[T1]{fontenc}    
\usepackage{hyperref}       
\usepackage{url}            
\usepackage{booktabs}       
\usepackage{amsfonts}       
\usepackage{nicefrac}       
\usepackage{microtype}      
\usepackage{xcolor}         
\usepackage{graphicx}
\usepackage{array}
\usepackage{enumitem}
\usepackage{graphicx}
\usepackage{amsmath}
\usepackage{mathtools} 
\usepackage{booktabs}
\usepackage{multirow}
\usepackage{wrapfig}
\usepackage[symbol]{footmisc}
\usepackage{caption} 
\title{Speech Foundation Models Generalize to Time Series Tasks from Wearable Sensor Data} 

\author{
Jaya Narain  \\
Apple \\
\texttt{jnarain@apple.com} 
\And
Zakaria Aldeneh\\
Apple \\
\texttt{zaldeneh@apple.com} \\
\And
Shirley Ren\\
Apple \\
\texttt{shirleyr@apple.com} \\
\And
}

\begin{document}

\maketitle

\begin{abstract}
Both speech and sensor time series data encode information in both the time- and frequency- domains, like spectral powers and waveform shapelets. We show that speech foundation models learn representations that generalize beyond the speech domain and achieve state-of-the-art performance on diverse time-series tasks from wearable sensors. Probes trained on features extracted from HuBERT and wav2vec 2.0 outperform those extracted from self-supervised models trained directly on modality-specific datasets for mood classification, arrhythmia detection, and activity classification tasks. We find that the convolutional feature encoders of speech models are particularly relevant for wearable sensor applications. The proposed approach enhances performance on data-scarce time-series tasks using simple probing methods. This work takes a step toward developing generalized time-series models that unify speech and sensor modalities.

\end{abstract}

\section{Introduction and Related Work}

Time series models have been trained for applications spanning numerous domains---including health, activity recognition, gesture recognition, weather forecasting, and infrastructure modeling.  Classical time series modeling methods include dynamic time warping, shapelet-based methods, convolution-based methods like ROCKET, and numerous other approaches (\cite{bagnall2017great, middlehurst2024bake}).  Recent work has also explored deep learning-based methods and representation learning strategies to enable shared learning among tasks (\cite{xu2023rebar, xu2024relcon, abbaspourazad2023large, erturk2025beyond}).  In the realm of physiological and wearable sensor data, most approaches have focused on within-domain representation learning where there is a domain match between pre-training data and evaluation tasks.   Prior works on transfer learning for time series modeling often focus on time series data like power consumption, traffic, and weather and often on forecasting tasks (\cite{woo2024unified, liu2023itransformer, jin2023time}). 

Voice2Series (\cite{yang2021voice2series}) explored re-programming speech processing models for time series tasks using task-specific target data along with a transformer-based speech model, and found strong performance across the evaluated UCR datasets---including some sensor-based tasks like ECG modeling.  Voice2Series trained custom speech embedding models from scratch, re-trained layers within the model for each time series task, and then identified source-to-target label maps for inference in the new domains.  While successful, this approach required both re-training and label mapping, as well as from-scratch training of speech foundation models. We use publicly available frozen pre-trained models directly as feature extractors, and train only lightweight probes for each task, without any re-training the embedding model backbone or label mapping. Similar between domain knowledge transfer has been successful in other areas---for instance, the Audio Spectrogram Transformer (\cite{gong2021ast}) showed improved performance on speech tasks by training a ViT model on ImageNet data.  

\begin{figure}[t!]
  \centering
  \includegraphics[width=\linewidth]{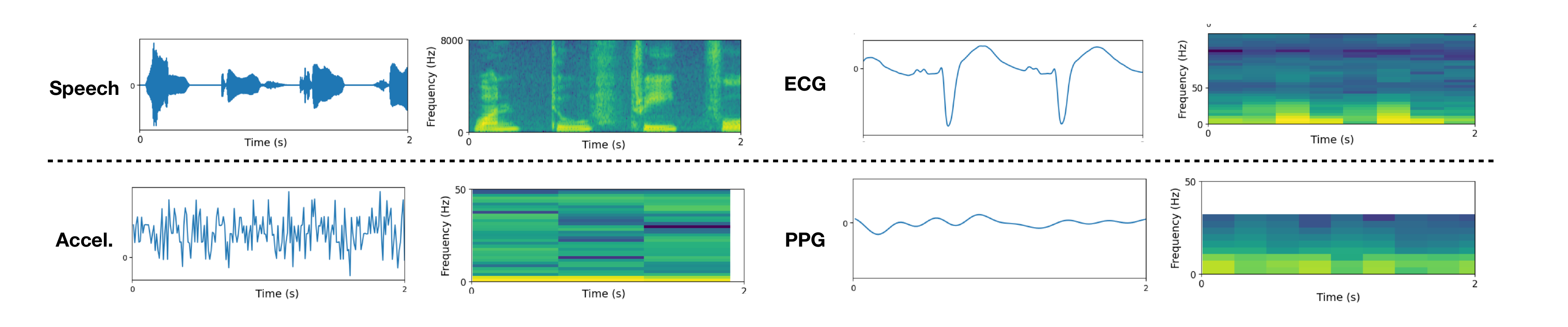}
  \caption{Time-series data from modalities such as speech, accelerometer (Accel.), electrocardiogram (ECG), and photoplethysmogram (PPG) signals contain rich temporal and spectral characteristics, including frequency band powers, periodic patterns, and distinctive waveform shapelets.}
  \label{fig:time_series}
\end{figure}

Speech and wearable data streams have related key signal properties: including frequency band powers, periodic structures, and shapelets in the time domain (see Figure~\ref{fig:time_series}).  Many sensor domains have limited data availability---learning relevant structure from speech data can have high impact on improving performance on data-scarce time series tasks.  Additionally, using a single embedding model across time series modalities can improve computational efficiency in multi-modal systems by enabling the deployment of a single model with task-specific adapters across modalities. 

Here, we explore the lightweight adaptation of pre-trained state-of-the-art foundation models like HuBERT (Large) and wav2vec 2.0 (Large) via probing.  To our knowledge, we are the first to directly use pre-trained speech models as feature extractors in the time series domain.  We envision this as a step towards more efficient cross-modality adaptation and generalized time series models that can leverage data-hungry architectures and learning from large-scale cross-modality datasets including speech and audio.

\section{Methods and Results}

We use pre-trained speech models as feature extractors for a variety of time series tasks.  We train probes on four tasks using other sensor modalities: activity classification from accelerometer data from a range of datasets and device placements, arrhythmia detection from ECG data, and stress classification using PPG.  For each task, we train linear probes and MLPs.  We report performance with an MLP probe at the first layer in each results table, along with per-layer performance across the transformer module for each task for both HuBERT and wav2vec 2.0. 
See the Appendix for additional details on each dataset and evaluation scheme.

\begin{figure}[b!]
  \centering
  \includegraphics[width=\linewidth]{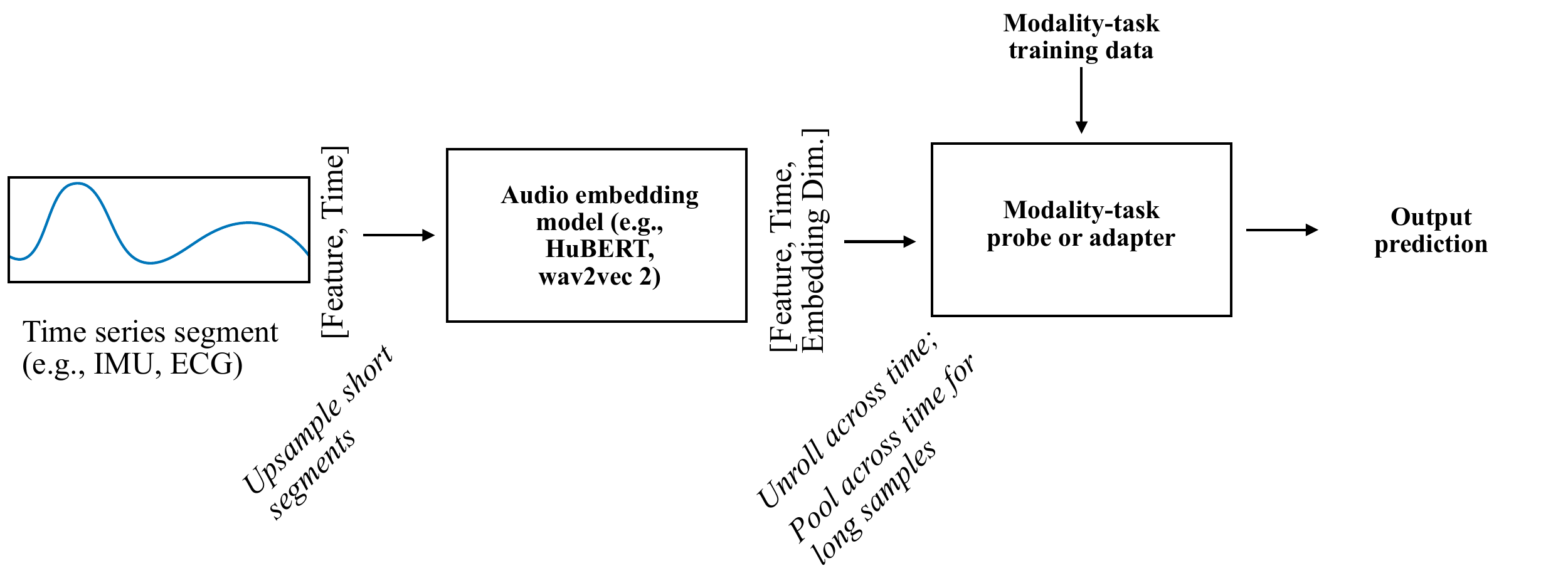}
  \caption{Speech foundation models as feature extractors for other modalities.  Time series data is fed as inputs into audio embedding models, with short segments upsampled.  Task specific probes are trained on the extracted features, and used to generate predictions across time series tasks.}
  \label{fig:feature_extraction_pipeline}
\end{figure}

\textbf{Activity Classification} We evaluated activity classification on four 3-axis accelerometer data datasets spanning three sensor positions and two window sizes.  In both evaluations, we included a benchmark using a Random Forest and engineered features, as in \cite{yuan2024self}.  We included results and benchmarks from two window sizes along with performance comparisons from \cite{xu2024relcon, yuan2024self, haresamudram2022assessing}: (1) following \cite{yuan2024self}, a leave one subject out subject out evaluation scheme with 10 second windows at 100 Hz with the PAMAP2 wrist data \cite{reiss2012introducing} (8 classes, \textit{n}=2,869) and the Opportunity dataset (4 classes, \textit{n}=3,842) \cite{roggen2010collecting}, and (2) following \cite{haresamudram2022assessing}, a 5-fold cross-validation evaluation on 2 second windows of 100 Hz 3-axis accelerometer data: HHAR \cite{reyes2015smartphone} wrist data (6 classes, \textit{n}=3,370), Motionsense waist data (6 classes, \textit{n}=14,121), \cite{Malekzadeh:2018:PSD:3195258.3195260}, and PAMAP2 leg data (12 classes, \textit{n}=9,709).  The 2 second windows were upsampled by a factor of 2 to have sufficient sample lengths to use with the pre-trained speech models.  The embeddings extracted from 10 second windows were pooled across the time dimension before being passed to the probe, in order to reduce dimensionality.  Table \ref{tbl:activity_classification} shows results from five activity classification evaluations: two with 10 second windows following the evaluation in \cite{yuan2024self} and three with 2 second windows following the evaluation in \cite{haresamudram2022assessing}.

\textbf{Arrhythmia Detection (ECG)}  We conducted binary arrhythmia classification using the MIT-BIH dataset \cite{moody1983new}, using 10 second windows sampled at 250 Hz.  The extracted embeddings were pooled across the time dimension before being passed to the probe, in order to reduce dimensionality.  We compare to previous results on this dataset reported in \cite{xu2023rebar}, including a baseline supervised model and self-supervised models trained on ECG data.  Results are reported in Table \ref{tbl:ecg_ppg}.

\begin{table}[]
\centering
\resizebox{.8\textwidth}{!}{%
\begin{tabular}{@{}ll@{\hspace{0.15cm}}cccccc@{}}
\toprule
& & PAMAP2 & Opportunity & HHAR & Motionsense & PAMAP2\\ 
& & (Wrist) & (Wrist) & (Wrist) & (Waist) & (Leg) \\ \midrule
& & \multicolumn{2}{l}{\cite{yuan2024self} Eval  } &  \multicolumn{3}{l}
{\cite{haresamudram2022assessing} Eval }\\ \midrule
& & \multicolumn{5}{c}{F1 (macro avg.)} \\ \midrule
& Feat eng. + RF & 66.6 ± 9.3 & 33.4 ± 8.8 & 61.2 ± 10.2 & 77.7 ± 3.5 & \textbf{67.0 ± 10.1} \\ \cmidrule(l){1-7} 
 \multirow{4}{*}{\rotatebox[origin=r]{90}{\parbox{1cm}{\centering Pretrain \\accel}}} & MLP (Yuan'24) & 72.5 ± 5.4 & 57.0 ± 7.8 & -- & -- & -- \\
 & SimCLR + Linear (Hare'22) & --  & --  & 55.9 ± 1.8 & 83.9 ± 1.8 & 50.8 ± 3.0 \\
 & SimCLR + MLP (Hare'22)  & -- & -- & 58.6 ± 2.2 & 85.6 ± 2.5 & 60.2 ± 2.3 \\
 & RelCon + MLP & \textbf{85.4 ± 3.5} & \textbf{69.1 ± 8.3} & 57.6 ± 3.2 & 80.4 ± 0.7 & 54.0 ± 0.8 \\ \midrule
 \multirow{4}{*}{\rotatebox[origin=r]{90}{\parbox{1cm}{\centering Pretrain speech}}} & HuBERT + Linear &  71.3 ± 7.4  & 49.3 ± 8.2 & 65.6 ± 8.6 & 80.0 ± 3.7 & 54.5 ± 4.3 \\
 & HuBERT + MLP & 73.6 ± 7.0 & 50.3 ± 3.9 & 69.0 ± 11.1 & 93.1 ± 2.5 & 60.5 ± 5.9 \\
 & wav2vec 2.0 + Linear & 67.1 ± 7.0 & 47.3 ± 3.9 & 70.2 ± 3.2 & 89.1 ± 3.4 &52.1 ± 4.7 \\
 & wav2vec 2.0 + MLP & 68.5 ± 6.0 & 50.8 ± 3.4 & \textbf{74.5 ± 3.7} & \textbf{93.4 ± 2.3} & 54.5 ± 6.2 \\ 
 \bottomrule
\end{tabular}%
}

\caption{\textbf{Activity classification results}. Probes with pre-trained speech models compared to probes with pre-trained IMU models from \cite{yuan2024self} "(Yuan'24)" and \cite{haresamudram2022assessing} "(Hare'22)" and a baseline Random forest model with engineered features.}
\label{tbl:activity_classification}
\vspace{-2mm}
\end{table}

\textbf{Mood Classification (PPG)} We conducted four-class mood classification (baseline, stress, amusement, and meditation) using PPG data from WESAD \cite{schmidt2018introducing}, in one minute windows sample at 64 Hz.  Results are reported in Table \ref{tbl:ecg_ppg}, along with comparative results with identical evaluation from \cite{xu2023rebar} including a baseline supervised model and self-supervised models trained on PPG data.




\begin{table}[b!]
\centering
\resizebox{.8\textwidth}{!}{%
\begin{tabular}{@{}ll@{\hspace{0.15cm}}cccc@{}}
\toprule
& & \multicolumn{2}{c}{Arrhythmia Detection (ECG)} &  \multicolumn{2}{c}
{Mood Classification (PPG)}\\ 
& & \multicolumn{2}{c}{MIT-BIH} &  \multicolumn{2}{c}
{WESAD}\\ \midrule
& & AUC & Accuracy & AUC & Accuracy\\  \midrule

&  NN (Xu'23) & 0.93 & 78.1 & 0.62 & 41.4  \\ \cmidrule(l){1-6} 
 \multirow{3}{*}{\rotatebox[origin=r]{90}{\parbox{0.25cm}{Pre-train ECG/ PPG}}} & SimCLR (Xu'23) & 0.83 & 69.9 & 0.62 & 34.5 \\
 & REBAR (Xu'23)  & 0.92 & 81.5 & 0.70 & 41.4 \\ \midrule
 \multirow{4}{*}{\rotatebox[origin=r]{90}{\parbox{1cm}{\centering Pretrain speech}}} & HuBERT + Linear &  0.89  & 87.0 & 0.79 & 57.3 \\
 & HuBERT + MLP & 0.95  & 94.0 & \textbf{0.82} & \textbf{77.5} \\
 & wav2vec 2.0 + Linear & 0.96 & \textbf{96.7} & 0.72 & 52.8 \\
 & wav2vec 2.0 + MLP & \textbf{0.97}  & 96.1 & 0.80 & 70.8 \\ 

 \bottomrule
\end{tabular}%
}

\caption{\textbf{Arrhythmia detection results and mood classification results}. Probes with pre-trained speech models compared to probes with pre-trained ECG models and pre-trained PPG models from \cite{xu2023rebar}.}
\label{tbl:ecg_ppg}
\vspace{-2mm}
\end{table}

\section{Discussion}
\textbf{Model performance} Probes trained on pre-trained speech foundation models had the best or competitive performance across tasks when using MLP probes trained on embeddings extracted from early layers.  Pre-trained speech representations outperformed both baselines and self-supervised models trained directly on sensor data for most tasks---activity classification with two second windows, mood classification, and arrhythmia detection.  Probes trained with pre-trained speech representations had competitive performance for activity classification with 10 second windows, though lower scores than the RelCon model trained on accelerometer data.  The longer windows may have had enough information to better leverage the learned motifs in RelCon -- better understanding how modeling strategy and sample window intersect will be explored in the future.

\begin{figure}[t!]
  \centering
  \includegraphics[width=\linewidth]{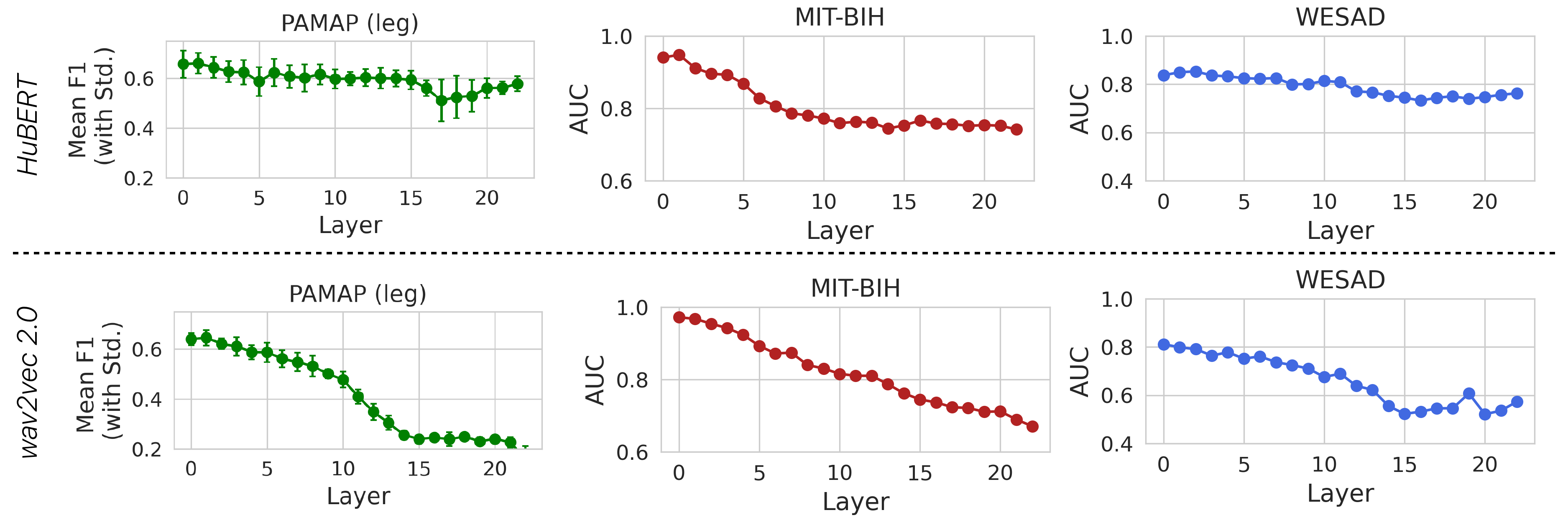}
  \caption{Performance by transformer layer with MLP probes for each task: activity classification (results shown with the PAMAP2 leg data), arrhythmia detection, and mood classification.  Early layer performance is better across modalities, particularly for wav2vec 2.0.}
  \label{fig:per_layer}
\end{figure}

\begin{figure}[b!]
  \centering
  \includegraphics[width=\linewidth]{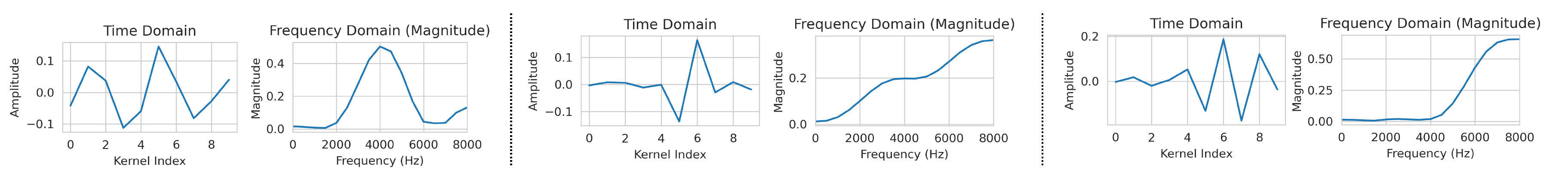}
  \caption{Visualization of selection of convolutional filters from HuBERT, from the first convolutional layer in the model.  The filters capture periodic and spiked shapelets, and include filters like bandpass filters and high-pass filters.}
  \label{fig:conv_viz}
\end{figure}

Earlier layer representations from the transformer module were consistently better than later layer representations, especially for wav2vec 2.0 (Figure \ref{fig:per_layer}).  This suggests that the convolutional feature extractor layers preceding the transformer module learned by the speech encoders are particularly relevant across domains.  Figure \ref{fig:conv_viz} shows sample convolution filters from HuBERT, which selected for visualization because they had high L2 norms and distinct properties.  The filters capture periodic and spiked shapelets, and include filters like bandpass filters and high-pass filters.  Additional work will further explore the interpretability of these filters across-domains.  

\textbf{Limitations} While speech and the investigated sensor domains (accelerometer data, ECG, and PPG) share enough commonalities to enable shared enocoders, the investigated sensor data streams were sampled at lower frequencies (100 Hz, 250 Hz, and 64 Hz respectively) than speech data (typically at 16,000 Hz with these models).  Future experiments will include ablations to assess impact of sampling rates, input processing, and window lengths as well as training strategies to better enable learning across diverse sample frequencies.  We present, to our knowledge, a set of first experiments showing that speech models generalize to wearable time series tasks -- additional validation is needed to further develop and understand how this methodology might be deployed, like evaluation with additional datasets, probing modeling biases, and further investigating interpretability. 

\textbf{Conclusions} The analyses show that pre-trained speech models can be effective feature extractors for wearable sensor tasks. Models like HuBERT and wav2vec 2.0 are trained with large, well-curated speech datasets, learning of rich and transferable representations. In contrast, wearable sensor datasets are typically smaller and task-specific. The convolution encoders from speech models were particularly impactful, suggesting that motif-learning in time series models may be particularly impactful.  Our results show that cross-domain learning---leveraging speech data---has strong performance on data-scarce tasks involving wearable sensors, suggesting cross-modality adaption and learning as an impactful modeling strategy.




\
\bibliographystyle{unsrtnat}
\bibliography{ref}

\section{Appendix}

\subsection{Datasets}

\textbf{PAMAP2 (Wrist)}
We extracted 10 second segments at 100 Hz from PAMAP2 \cite{reiss2012introducing}, and used a leave one subject out evaluation scheme following the procedure in \cite{yuan2024self}.  The evaluation included eight classes: lying, sitting, standing, walking, ascending stairs, descending stairs, vacuum cleaning, and ironing.  The experiments included \textit{n}=2,860 samples from eight participants.

\textbf{Opportunity (Wrist)}
We extracted 10 second segments at 100 Hz from \cite{roggen2010collecting}, and used a leave one subject out evaluation scheme following the procedure in \cite{yuan2024self}.  The evaluation included four classes: sitting, standing, walking, lying down.  The experiments included \textit{n}=3,842 samples from four participants.

\textbf{HHAR (Wrist)}
We extracted 2 second segments at 100 Hz from \cite{reyes2015smartphone}, and used a 5-fold cross validation evaluation scheme using the procedure in \cite{haresamudram2022assessing}.  The evaluation included six classes: stairs down, stairs up, walk, bike, stand, sit.  The experiments included \textit{n}=3,370 samples.

\textbf{Motionsense (Waist)}
We extracted 2 second segments at 100 Hz from \cite{Malekzadeh:2018:PSD:3195258.3195260}, and used a 5-fold cross validation evaluation scheme using the procedure in \cite{haresamudram2022assessing}.  The evaluation included six classes: stairs down, stairs up, walk, jog, stand, sit.  The experiments included \textit{n}=14,121 samples.

\textbf{PAMAP2 (Leg)}
We extracted 2 second segments at 100 Hz from \cite{reiss2012introducing}, and used a 5-fold cross validation evaluation scheme using the procedure in \cite{haresamudram2022assessing}.  The evaluation included twelve classes: rope jumping, lying, sitting, standing, walking, running, cycling, Nordic walking, ascending stairs, descending stairs, vacuum cleaning, ironing.  The experiments included \textit{n}=9,709 samples.

\textbf{MIT-BIH}
We extracted 10 second segments at 250 Hz from \cite{moody1983new}.  We followed the pre-processing and evaluation process as described in \cite{xu2023rebar} to allow for comparison with prior results from self-supervised models trained in-domain.

\textbf{WESAD}
We extracted 60 second segments at 64 Hz from \cite{schmidt2018introducing}.  We followed the pre-processing and evaluation process as described in \cite{xu2023rebar} to allow for comparison with prior results from self-supervised models trained in-domain.

\end{document}